\documentclass[letterpaper,twocolumn,10pt]{article}
\usepackage{usenix,epsfig,endnotes, biblatex}
\begin{document}

\date{}

\title{\Large \bf Machine Learning Optimization for Enhanced OS Fingerprinting}

\author{
{\rm Spencer\ Ekeroth}\\
Virginia Tech
\and
{\rm Jeremy\ Neale}\\
Virginia Tech
\and
{\rm Jae Sung\ Kim}\\
Virginia Tech
}

\maketitle

\thispagestyle{empty}

\subsection*{Abstract}
\hspace{0.5 cm} Operating System (OS) Fingerprinting is a technique that can be used to identify a network's operating systems by evaluating network traffic in the form of TCP/IP packets. This research will explore the effectiveness of passively identifying operating systems on the CIC-IDS2017 dataset, a collection of over 47 gigabytes of pcap files with their corresponding operating systems. This research also proposes a new command line interface, OsirisML, which uses \textit{nPrint}\cite{nPrint2024} to preprocess the data into tabular data and \textit{XGBoost}\cite{XGBoost} to apply ML to the data to generate, retrain, and test ML models. When packets are split randomly between training and testing, OsirisML models reach an accuracy of 97.66\% on a down-sampled subset of the Friday capture and 84.69\% on the entire capture. On the entire Monday capture, which contains no attacks, OsirisML reaches an accuracy of 73.83\% and an F-1 score of 79.38\%.

\section{Introduction}

\hspace{0.5 cm}Passive fingerprinting analyzes network traffic offline, which avoids detection from the network. There are many versions of passive fingerprinting tools that can analyze packet capture (\textit{pcap}) data and attempt to identify the operating system. To generate a new or custom passive fingerprinting model using machine learning, \textit{pcap} data can be analyzed and used to generate a new model with a tool like \textit{nPrintML}, which uses \textit{nPrint} to transform the PCAP data into tabular data, and \textit{AutoML} \cite{AutoML} automatically generate a ML model from the tabular data.

Given the dynamic nature of network environments, where new OS versions are continuously released, there is a compelling need for tools that can adapt and retrain machine learning models for passive OS fingerprinting to remain effective against the ever-evolving landscape of operating systems.

This research introduces a novel command-line tool, OsirisML, which consists of two main components similar to \textit{nPrintML}. First, it leverages \textit{nPrint} to transform network traffic (pcap files) into a normalized tabular format suitable for machine learning. Unlike \textit{nPrint}s alternative, \textit{nPrintML}, that utilizes \textit{AutoML} for automated model generation, OsirisML integrates the advanced capabilities of XGBoost—an algorithm renowned for its efficiency with tabular data—to enhance model accuracy and adaptability. The usage of XGBoost \cite{XGBoost} for machine learning applications contains 4 components: generating models, testing unseen data with a model (passive fingerprinting), retraining a model on an unseen dataset to generate a new model, and hyperparameter tuning (section 9).

This tool allows for 3 significant advantages over its counterparts:

\begin{itemize}
    \item Hands-on manipulation of the model creation with custom hyperparameter tuning, file partitioning, and retraining models
    \item Improved accuracy and F-1 scores from model generation on identical data
    \item More efficient RAM usage on identical data
\end{itemize}

\section{Motivation and problem statement}

\hspace{0.5 cm} As aspiring cybersecurity professionals, it is understood that the importance of identifying an operating system over a network can be of critical importance. Creating an accurate tool to do so would be beneficial for independent work, as well as to the global community of white hat threat actors. It could be used to impersonate a black-hat threat actor to see if they could enumerate a network and exploit any vulnerabilities.

Open source work is what has propelled the computer science field forward for the past century. Projects such as Linux, GNU, Python, and Docker have benefited millions of people across the world. Improving upon pre-existing open source tools would be one way to advance the fields of cybersecurity and machine learning and help other computer scientists in the future.

The goal of OsirisML is not just to make an all-in-one passive fingerprinting model, but to provide a method to create and build upon pre-existing models. This allows OS Fingerprinting to stay up to date and continuously improve from open-source contributions of datasets and model generation.

\section{Related work}
\hspace{0.5 cm} In the research paper, 'New Directions in Automated Traffic Analysis'~\cite{Holland2021}, Jordan Holland, Paul Schmitt, Nick Feamster, and Prateek Mittal created a tool called \textit{nPrint} and \textit{nPrintML}~\cite{nPrint2024}. \textit{nPrint} is capable of capturing traffic live or processing a pcap file to convert raw packet data into a tabular format suitable for machine learning algorithms. \textit{nPrintML} is an all inclusive tool that takes the tabular output of \textit{nPrint} and uses \textit{AutoML} to select a machine learning algorithm to create an optimized model for OS fingerprinting. As more machine learning models are created and better optimized to handle tabular data, this project will focus on optimizing the \textit{nPrint} data output and using modern machine learning algorithms to attain a model that is more accurate in passive OS fingerprinting than what was achieved by Jordan Holland et al.~\cite{Holland2021} \\

\section{Production Environment for OsirisML}

\hspace{0.5 cm} The production environment for OsirisML is set up on a virtual machine running Ubuntu 22.04 \cite{Ubuntu2204}, a Debian-based Linux distribution, to ensure stability and consistency. The virtual machine is running on an AMD EPYC 7401 24-Core Processor with 128GB of RAM. Virginia Tech's rlogin cluster, which is a multi-machine system running Linux to provide computational support for Computer Science instruction, was also utilized to run \textit{hyperparameter\_tuning.py}. The rlogin cluster has many nodes, each equipped with 2 Xeon Gold 5218 CPUs with 2.3Ghz 16 core processors, 384 GB of RAM, and a 10 Gbit Ethernet connection \cite{VTClusterHardware2024}. The results in section 10 were reproduced on a workstation with 32 threads and 123 GB of RAM using \textit{XGBoost} 3.2.0.

\section{CIC-IDS 2017 PCAP Dataset}
\hspace{0.5 cm}The CIC-IDS 2017 dataset \cite{CIC-IDS2017}, released by the University of New Brunswick, is a comprehensive collection of network traffic designed to represent a mix of the typical network traffic in a real-world environment, along with intrusion attempts from Kali Linux. This dataset is chosen because it contains a large set of labeled operating systems by IP address. The networks are represented by an array of operating systems, including different versions of Ubuntu servers, Windows machines from Vista to Windows 10, and macOS systems, among others. Each operating system is labeled by the source IP address of a single host, or two hosts for the Ubuntu server and the Ubuntu web server, so in this dataset an operating system and a host are the same. The Friday capture contains 13 labeled operating systems, including the Kali Linux attacker, while the Monday capture, which contains no attacks, has 12. Operating system fingerprinting often relies on discerning the subtleties in packet structure, sequence numbers, fragments, window sizes, and other TCP/IP stack behaviors that vary from one OS to another.

\section{nPrint and nPrintML}

\hspace{0.5 cm}The starting point for developing a model that is capable of differentiating fine-grain operating systems was to replicate Section 5.2 in 'New Directions in Automated Traffic Analysis'~\cite{Holland2021}. By using the same features, it would provide a benchmark for the new model. \textit{nPrint} takes a unique approach to packet representation by assuming all packets have a header of maximum length. Any missing options or fields in the headers are represented with a value of -1. To keep consistency with Section 5.2 only IPv4 and TCP headers will be extracted from each packet. This results in each encoded packet sample to have 960 features, representing the 480 bits from the IPv4 header and the additional 480 bits from the TCP header.

The output generated from \textit{nPrint} with the IPv4 and TCP header options set still requires cleaning the IP source address, IP destination address, IP identification, TCP source port, TCP destination port, and both TCP sequence and ack number, which can act as direct identifiers to the operating system. OsirisML drops these 177 bit columns along with the source IP address column, which leaves each sample with 783 features. Without this cleaning, an \textit{XGBoost} model trained on the Monday capture reached an accuracy and F-1 score of 100\%, and its most important features were source IP, destination port, and sequence number bits. With the cleaning, the same model reached an accuracy of 63.36\%. These changes can be made to the source code of \textit{nPrintML}. After using pip to install the \textit{nPrintML} package, it is imperative to edit \textit{automl.py} which can be found in \textit{site-packages/nprintml/learn/} where site-packages is the directory pip installs new libraries and packages. Before the \textit{train} method is called, the areas of the IPv4 and TCP header mentioned above need to be dropped as they act as direct identifiers. If the areas are not dropped, \textit{nPrintML} will return a model that is extremely overfitted to the dataset due to data leakage.

Removing these fields does not remove every host-specific value. Some values are constant for a given machine. The TTL, window size, and TCP options are legitimate features of an operating system, but since each operating system in the dataset is a single host, they can also identify the host itself (section 12).

\begin{figure*}[p]
    \includegraphics[width=\textwidth]{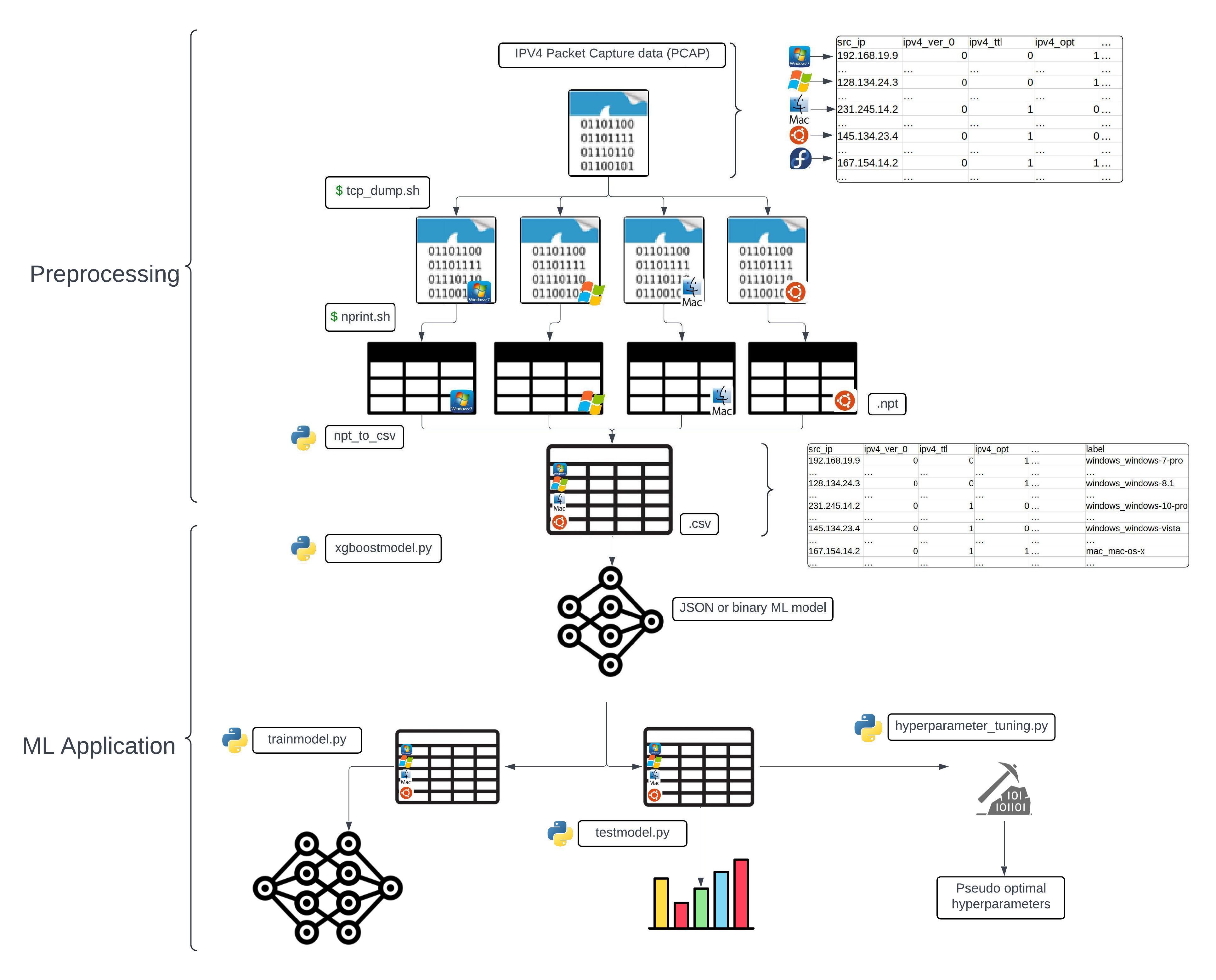}
    \caption{Visualization for the flow of both phases of the CLI}
    \label{fig:osirisml}
\end{figure*}

\section{Phase 1 - Data Preprocessing Workflow}

\hspace{0.5 cm}OsirisML is split into two main components: preprocessing, and ML application. The goal of the preprocessing is to take \textit{pcap} data and convert it to tabular (\textit{csv}) data, making it compatible for machine learning algorithms. This entire process can be executed with \texttt{make preprocess PCAP=<PCAP\_file\_name.pcap>}, which runs \texttt{process\_pcap.sh} and calls the following three scrips found in \textit{preprocessing/} in succession:
\hspace{0.5 cm}

\begin{enumerate}
    \item \textbf{tcp\_dump.sh} - uses the source IP labels provided on CIC-IDS2017 host website~\cite{CIC-IDS2017} by default to split up traffic based on operating systems. Alter the source code of this file with new source IPs if attempting to generate a model from a dataset with different source IPs.

    Note that the PCAP file that this script accepts as a system argument \textbf{must} be located in \textit{data/pcap}, and the resulting PCAP files will be placed into \textit{data/pcap/pcap\_os\_split/}. The name of each file will be the name of the operating system provided in the source code.

    \texttt{Usage: ./tcp\_dump.sh <PCAP\_file\_name.pcap>}

    \item \textbf{nprint.sh} - unlike \textit{nPrintML}, \textit{nPrint} is unable to take a directory of pcaps as a valid argument. This script loops through all PCAP files in \textit{data/pcap/pcap\_os\_split/} and calls \textit{nPrint} on each file with -4 -t parameters - see \textit{nPrint} documentation.

    The resulting NPT files will be placed into \textit{data/npt/} and will have the same names with new extensions.

    \texttt{Usage: ./nprint.sh}
    
    \item \textbf{npt\_to\_csv.py} - This file loops through each NPT file in \textit{data/npt/} and writes each line into a single CSV file, which it places in \textit{data/csv/}.

    \texttt{Usage: python3 <npt\_to\_csv.py> <new.csv>}
\end{enumerate}

Execute this entire workflow with:

\texttt{make preprocess PCAP=<PCAP\_file\_name.pcap>} \\

Note that process\_pcap.sh also deletes the \textit{data/npt/} and \textit{data/pcap/pcap\_os\_split/} directories after finishing.

\section{Phase 2 - ML Application with XGBoost}
\hspace{0.5 cm} Once the tabular data is preprocessed, it has the relevant TCP/IP binary input attributes and one classification label, which is the corresponding operating system. With the labeled CSV file, different ML algorithms can easily be applied to the data to produce a model. XGBoost was selected for its effectiveness with tabular data, along with its open-source nature and extensive documentation.

\hspace{0.5 cm} The XGBoost scripts splits the CSV file into \textit{x\_train, x\_test, y\_train, and y\_test}. The $x$ files are the input attributes, which in this case are the TCP/IP binary input attributes, and the $y$ files are the corresponding classification labels, which is the last column of the CSV. The labeled csv dataset has 962 columns (or attributes), but the first column is the source IP address in semantic form, and the last column is the classification label as a string.

\hspace{0.5 cm} The x\_train and x\_test values were then further cleaned to remove the bit ranges of values that would act as direct identifiers to the model. Both y\_train and y\_test were encoded from strings to integers to allow them to be processed by XGBoost's classifier.

\hspace{0.5 cm} The default eval metric supplied to all three model scripts is \textit{mlogloss}, which is selected due to its nature of multi-class classification, and its robustness to imbalanced classes.

There are 4 main files in \textit{model/}:

\begin{enumerate}
    \item \textbf{hyperparameter\_tuning.py} - see section 9.
    
    \texttt{Usage: python3 hyperparameter\_tuning.py <data.csv>}
    
    \item \textbf{xgboostmodel.py} - generates a model from the supplied \textit{csv}. This uses a test size of 0.2 by default, but can be updated with an optional sys argument.

    The \textit{csv} file must be located in \textit{data/csv/}, where it should be by default after data preprocessing. The outputted \textit{json} file will be placed into \textit{model/json/} by default.

    \texttt{Usage: python3 xgboostmodel.py <data.csv> <modelname> <OPTIONAL: <decimal for test split size>}
    
    \item \textbf{trainmodel.py} - \textit{XGBoost} allows for already generated models to be retrained on new data to produce a new model. This is useful for optimized RAM and storage, as well as improving the robustness of a model. This script uses a \textit{DMatrix}, where the $x$ and $y$ train and assigned to \textit{dtrain}, and $x$ and $y$ test are assigned to \textit{dval}. This matrix approach during training drastically improves space and speed.

    \hspace{0.5 cm} When invoking \textit{xgboost.train}, the loaded model is used as the starting point, the max number of additional boost rounds is set to 1000, and early stopping rounds is set to 10, using 25\% of the training data for validation. The model is then saved at its best round. The new \textit{csv} must contain the same operating systems as the original model, since the labels are encoded in the same order. Currently, max\_depth = 10 is the only hyperparameter supplied to this script.

    \hspace{0.5 cm} The model must already be located in \textit{model/json/}, which is should be after invoking xgboostmodel.py or this script. The training data must be located in \textit{data/csv/}. The new model will be placed into \textit{model/json/}.

    \texttt{Usage: python3 trainmodel.py <current\_model.json> <data.csv> <new.json>}
    
    \item \textbf{testmodel.py} - this script can be thought of as passive OS Fingerprinting. It takes a model generated by xgboostmodel.py or trainmodel.py, and tests unseen data to evaluate its performance. This script currently only returns the accuracy and F-1 score to evaluate the success of the model, but does not provide information on each individual machine in the packet capture data. To determine predictions for individual machine's operating system, supply only 1 source IP address in the \textit{csv} file to the model. The testing \textit{csv} must contain the same operating systems as the training \textit{csv}, since the labels are encoded from the testing file.

    The model must already be located in \textit{model/json/}, which is should be after scripts 2 and 3. The testing data must be located in \textit{data/csv/}.

    \texttt{Usage: python3 testmodel.py <model.json> <data.csv>}
\end{enumerate}

\section{Hyperparameter tuning and optimization}

\hspace{0.5cm} In machine learning, \textit{parameters} are learned and updated throughout the training process, while \textit{hyperparameters} are set beforehand and determine key features of the training method such as model architecture, learning rate, and model complexity. \textit{XGBoost} uses a gradient boosted decision tree machine learning algorithm that accepts many optional hyperparameters. Using optimal hyperparameters for the machine learning training process is crucial to achieving higher model accuracy and F-1 scores while avoiding overfitting. Hyperparameters also have a significant role in the time and computer resources necessary to train a model.

\hspace{0.5cm} OsirisML comes equipped with a script $\textit{model/hyperparameter\_tuning.py}$ that searches for the optimal parameters for the supplied CSV file, which is the PCAP data that has been preprocessed into tabular data. See section 8 for usage.

\hspace{0.5cm} To determine the optimal parameters, the script uses $\textit{GridSearchCV}$ \cite{scikit-learn}, which generates an XGBoost model for every combination of hyperparameters supplied as a parameter of a Python Dictionary, where the keys are the hyperparameter options as strings, and the values are the options for that hyperparameter. For example,

\begin{verbatim}
    # Create the parameter grid
    param_grid =
    {
        'max_depth': [3, 6, None],
        'gamma': [0.1, 0.5, 1.0],
        'min_child_weight': [2, None]
    }
\end{verbatim}

\hspace{0.5cm} Using GridSeachCV with this grid would generate $18$ models, since there are $3 * 3 * 2$ combinations of the three parameters. After generating the $18$ models, GridSeachCV returns the best combination of parameters based on the supplied evaluation metric, which is supplied as accuracy in the script, using 5-fold cross validation on the training data.

\hspace{0.5cm} It is recommended to use a smaller partition of the data to search for hyperparameters, then to use these psuedo-optimal hyperparameters on the entire dataset. This is because searching for hyperparameters is computationally expensive and costs lots of time and computer resources. The \textit{pcap} files can be partitioned with \texttt{python3 preprocessing/shrink\_pcap.py <path\_to\_pcap.pcap> ---fraction X (default 10)}.

\hspace{0.5cm} After retrieving the optimal combination of supplied hyperparameters, they can be supplied to $\textit{model/xgboostmodel.py}$ and used to train the entire data set.

\hspace{0.5cm} Using the $\textit{Friday-WorkingHours.pcap}$ file from the section 5 dataset, a partition of the dataset was created that split the entire 8.2 gb dataset $32$ times. Windows 10 and the Ubuntu web server were both split 50 times, since they were drastically over represented compared to the other OSes. The resulting dataset had 134,269 samples and the \textit{csv} file was now about 336 mb. The OSes were distributed by the following chart:

\noindent\includegraphics[width=\columnwidth]{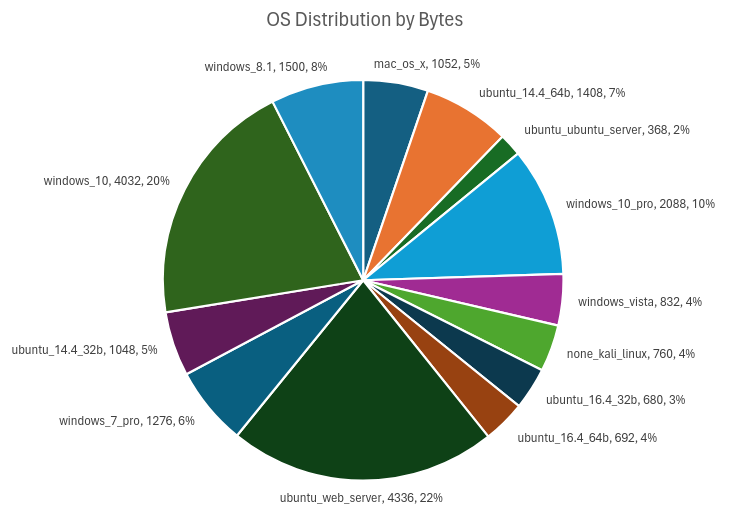}

With this smaller dataset, the hyperparameter tuning script reported that max\_depth = 10 and the others as \textit{None} was optimal, yielding the highest accuracy and F-1 scores of $97.70\%$ and $97.51\%$ respectively. There are many more hyperparameters to test, but due to the nature of the time it takes to generate models, only a select pool of hyperparameters were given to $\textit{GridSearchCV}$:

\begin{verbatim}
    # 54 combinations - 3 * 3 * 2 * 3
    param_grid =
    {
        'n_estimators': [400, 800, None],
        'max_depth': [6, 10, None],
        'min_child_weight': [2, None]
        'gamma': [0.1, 0.5, 1.0]
    }
\end{verbatim}

Further research could test more hyperparameters and/or use a larger dataset to return a more accurate assessment of the optimal hyperparameters to generate models with.

\hspace{0.5cm} When training on the entire Friday file, the default learning rate of 0.3 occasionally caused \textit{XGBoost} to diverge. In one run the validation \textit{mlogloss} fell to 0.54 after 30 rounds and rose to 9.66 by round 50, and the final model was much less accurate. This does not happen on every run, but to prevent it, all results in section 10 use early stopping with 10 rounds of patience on 10\% of the training data.

\section{Results}

\hspace{0.5 cm} All results in this section use the 783 features left after the cleaning in section 6, max\_depth = 10 unless noted, and early stopping as described in section 9. All F-1 scores are macro averages.

\hspace{0.5 cm} The original results were first reproduced using a random 80\% training and 20\% testing split of the packets (Table~\ref{tab:reproduced}). Using the condensed version of \textit{Friday-WorkingHours.pcap} from section 9 produced an accuracy of 97.66\% and an F-1 score of 97.35\%, while \textit{nPrintML} reported 96\% accuracy and 96\% F-1 score on a random 70\% training and 30\% testing split of the same packets (\textit{nPrintML} only reports results with two digits). Using \textit{XGBoost}'s default parameters on the entire Friday file resulted in 78.40\% accuracy and 73.67\% F-1 score. With max\_depth = 10, the accuracy and F-1 score improved to 84.69\% and 82.61\% respectively. When using \textit{nPrintML} on the same dataset, the process killed due to RAM usage. The entire Friday file is an 8.2 gb \textit{pcap} file that was preprocessed into an 11.2 gb \textit{csv} file with 4,501,929 samples.

\hspace{0.5 cm} The final test was to replicate Section 5.2 of Jordan Holland et al.~\cite{Holland2021} with the first 100,000 packets of each operating system in \textit{Friday-WorkingHours.pcap}. The Ubuntu server and Kali Linux have fewer than 100,000 packets, so the dataset is comprised of 1,119,205 packets. Using this dataset saw an accuracy of 89.83\% and an F-1 score of 91.15\%.

\begin{table}[h]
\centering
\small
\begin{tabular}{lcc}
\hline
Dataset & Reported & Reproduced \\
\hline
Friday subset & 97.70 / 97.51 & 97.66 / 97.35 \\
Friday, default & 74.09 / 64.83 & 78.40 / 73.67 \\
Friday, max\_depth = 10 & 84.91 / 82.96 & 84.69 / 82.61 \\
Section 5.2 & 90.11 / 91.39 & 89.83 / 91.15 \\
Monday & -- & 73.83 / 79.38 \\
\hline
\end{tabular}
\caption{Reported and reproduced accuracy / F-1 score (\%), random split}
\label{tab:reproduced}
\end{table}

\hspace{0.5 cm} On the entire Friday file with max\_depth = 10, every operating system has an F-1 score between 87.83\% and 100\% except Windows 10 Pro (51.53\%), Windows 7 Pro (45.07\%), and Windows 8.1 (52.01\%), which are mostly confused with each other and with Windows 10.

\hspace{0.5 cm} To test OsirisML on a capture with no attack traffic, the same random split was used on the entire \textit{Monday-WorkingHours.pcap} file, which contains 5,387,586 packets from 12 operating systems. This produced an accuracy of 73.83\% and an F-1 score of 79.38\%. macOS, the Ubuntu web server, and Windows Vista have the highest F-1 scores (98.56\%, 95.02\%, and 87.81\%), while Windows 7 Pro (48.29\%) and Windows 10 Pro (57.14\%) have the lowest.

\hspace{0.5 cm} \textit{nPrintML} also evaluates its models on a random split of packets~\cite{nPrint2024}, so its reported results are directly comparable with the random split results of OsirisML. The \textit{nPrintML} results were not re-run for this revision. \textit{nPrintML} requires Python 3.8 and its own AutoML environment, and on the entire Friday file it was killed due to RAM usage.

\section{Proposed work}

\hspace{0.5 cm} Looking ahead with OsirisML, the plan is to train the model on the rest of the \textit{pcap} files within the CIC-IDS 2017 dataset and to test it on a day that was not used for training. Given enough memory, \textit{nPrintML} should also be run on the entire Friday file so that the two tools can be compared on the full dataset.

Another direction is a hierarchical model, which would first predict the operating system family and then predict the version within that family with a separate model. Feature importance on the Monday capture shows that the TTL separates the operating system families, while the TCP options separate the Ubuntu versions and the window size separates the Windows versions. Windows 10 and Windows 10 Pro use the same network stack, so they could be combined into one label.

OsirisML is an open-source project, and users will be able to contribute to the model by providing new \textit{pcap} data, if approved by the OsirisML team. A new dataset could be produced by isolating a network and populating it with several machines for each operating system, which would allow complete control over the collection of data and separate the operating system from the host.



\section{Potential Limitations}

\hspace{0.5 cm}While OsirisML reproduces its original results, there are some factors that limit how they should be interpreted. Each operating system in CIC-IDS2017 is a single host, or two hosts for the servers, so the model may learn the behavior of a host rather than of an operating system. The dataset is also imbalanced: Windows 10 has 1,595,579 packets on Friday while Kali Linux has 8,991, so the F-1 score should be reported along with the accuracy. OsirisML has only been tested on the CIC-IDS2017 dataset which only consists of 13 operating systems. In order to be confident in the success of this CLI, its success would need to be replicated with other datasets.

Current compatibility limitations are that OsirisML is currently only available for installation on Debian-based Linux. This is because it is recommended to use this tool in a VM, which is likely GNU-Linux based. Containerizing could make the tool more accessible and consistent across host machines.

\printbibliography

@InProceedings{Holland2021,
  author =       "Jordan Holland and others",
  title =        "{New Directions in Automated Traffic Analysis}",

  booktitle =    "Proceedings of the 2021 ACM SIGSAC Conference on Computer and Communications Security",
  year =         "2021",
  pages =        "3366--3383",
  DOI =          "https://doi.org/10.1145/3460120.3484758"
}

@misc{nPrint2024,
  author = {{The nPrint Project}},
  title = {nPrint},
  year = {2024},
  howpublished = {\url{https://nprint.github.io/nprint/}},
  note = {Accessed: 2024-02-27}
}

@inproceedings{CIC-IDS2017,
  title        = {Toward Generating a New Intrusion Detection Dataset and Intrusion Traffic Characterization},
  author       = {Sharafaldin, Iman and Lashkari, Arash Habibi and Ghorbani, Ali A.},
  booktitle    = {Proceedings of the 4th International Conference on Information Systems Security and Privacy (ICISSP)},
  year         = {2018},
  address      = {Portugal},
  month        = {January},
  organization = {Canadian Institute for Cybersecurity, University of New Brunswick}
}

@Misc{XGBoost,
  author = {{XGBoost Development Team}},
  title = {XGBoost: Scalable Machine Learning System},
  year = {2024},
  howpublished = {\url{https://xgboost.readthedocs.io/en/stable/}},
  note = {Accessed: 2024-05-03}
}

@Misc{AutoML,
  author = {AutoML Team},
  title = {AutoML: Automated Machine Learning},
  year = {2024},
  howpublished = {\url{https://www.automl.org/automl/}},
  note = {Accessed: 2024-05-03}
}

@Misc{scikit-learn,
  author = {{scikit-learn developers}},
  title = {GridSearchCV -- scikit-learn 1.1.3 documentation},
  howpublished = {\url{https://scikit-learn.org/stable/modules/generated/sklearn.model_selection.GridSearchCV.html}},
  note = {Accessed: 2024-05-01},
  year = {2023},
  key = {GridSearchCV}
}

@Misc{Ubuntu2204,
  title = {Ubuntu 22.04.4 LTS (Jammy Jellyfish)},
  howpublished = {\url{https://releases.ubuntu.com/jammy/}},
  note = {Accessed: 2024-04-12},
  year = {2024}
}

@Misc{VTClusterHardware2024,
  author = {Virginia Tech},
  title = {Hardware Information for Cluster},
  howpublished = {\url{https://wordpress.cs.vt.edu/rlogin/more-information-3/}},
  note = {Accessed: 2024-04-02},
  year = {2024}
}

\end{document}